\documentclass{article}

\PassOptionsToPackage{numbers, compress}{natbib}

\usepackage[final]{neurips_2023}

\usepackage[utf8]{inputenc}
\usepackage[T1]{fontenc}
\usepackage[usenames,dvipsnames,svgnames,table]{xcolor}
\usepackage[colorlinks,allcolors=Blue]{hyperref}
\usepackage{url}
\usepackage{booktabs}
\usepackage{amsfonts}
\usepackage{nicefrac}
\usepackage[activate=true,
final,
tracking=false,
kerning=true,
spacing=true,
factor=1050,
stretch=30,
shrink=30]{microtype}
\usepackage{enumitem}
\usepackage{xcolor}
\definecolor{darkpurple}{RGB}{75,0,130}
\definecolor{myblue}{RGB}{56,140,229}
\definecolor{mybrown}{RGB}{93,78,52}

\definecolor{prefixcolor}{RGB}{226,108,36}
\definecolor{xysuffix}{RGB}{0,104,220}
\definecolor{asuffix}{RGB}{178,39,238}
\usepackage{mathtools}
\usepackage{amsmath}
\usepackage{amsthm}
\usepackage{graphicx}
\usepackage{wrapfig}
\usepackage{subcaption}
\usepackage{bbm}
\usepackage{bm}
\usepackage{multirow}
\usepackage{booktabs}
 
\renewcommand{\vec}[1]{{\boldsymbol{\mathbf{#1}}}} 
\newcommand{\vh}{\vec{h}}
\DeclareMathOperator*{\argmax}{arg\,max}

\newcommand{\defeq}{\mathrel{\stackrel{\textnormal{\tiny def}}{=}}} 
\newcommand{\pluseq}{\mathrel{+\!\!=}}

\usepackage{algorithm}
\usepackage[endLComment=]{algpseudocodex}
\renewcommand\algorithmicdo{:}
\renewcommand\algorithmicthen{:}

\usepackage{xspace}
\makeatletter
\newcommand{\smartperiod}{\@ifnextchar.{}{.\@\xspace}}
\newcommand{\smartcomma}{\@ifnextchar.{}{,}\xspace}
\makeatother
\newcommand{\latin}[1]{#1}
\newcommand{\eg}{\latin{e.g.}\smartcomma}

\newcommand{\etc}{\latin{etc}\smartperiod}

\newcommand{\xTy}{\bm x\circ\mathcal{T}\circ\bm y}
\newcommand{\Txy}{\mathcal{T}_{\bm x,\bm y}}
\newcommand{\prefix}{_\text{prefix}}
\newcommand{\suffix}{_\text{suffix}}

\usepackage[capitalize]{cleveref}
\crefname{page}{page}{pages}
\crefname{footnote}{footnote}{footnotes}
\crefname{equation}{equation}{equations}
\crefname{corollary}{Corollary}{Corollaries}
\crefname{line}{line}{lines}
\crefrangeformat{line}{lines #3#1#4--#5#2#6}
\crefname{lstlsting}{Listing}{Listings}   
\crefname{section}{\S}{\S\S}
\Crefname{section}{\S}{\S\S}
\crefformat{section}{#2\S#1#3}
\Crefformat{section}{#2\S#1#3}
\crefrangeformat{section}{\S\S#3#1#4--#5#2#6}
\Crefrangeformat{section}{\S\S#3#1#4--#5#2#6}
\crefmultiformat{section}{\S#2#1#3}{ and~\S#2#1#3}{, \S#2#1#3}{ and~\S#2#1#3}
\Crefmultiformat{section}{\S#2#1#3}{ and~\S#2#1#3}{, \S#2#1#3}{ and~\S#2#1#3}
\crefrangemultiformat{section}{\S\S#3#1#4--#5#2#6}{ and~\S\S#3#1#4--#5#2#6}{, \S\S#3#1#4--#5#2#6}{ and~\S\S#3#1#4--#5#2#6}
\Crefrangemultiformat{section}{\S\S#3#1#4--#5#2#6}{ and~\S\S#3#1#4--#5#2#6}{, \S\S#3#1#4--#5#2#6}{ and~\S\S#3#1#4--#5#2#6}

\newcommand{\scorer}{\tilde{p}_{\theta}}

\usepackage[usenames,dvipsnames,svgnames,table]{xcolor}
\usepackage[]{todonotes}
\makeatletter
\newcommand*\iftodonotes{\if@todonotes@disabled\expandafter\@secondoftwo\else\expandafter\@firstoftwo\fi}
\makeatother
\newcommand{\noindentaftertodo}{\iftodonotes{\noindent}{}\ignorespaces}

\newcommand{\Fixme}[2][]{\noindentaftertodo}
\newcommand{\Notewho}[3][]{\noindentaftertodo}
\newcommand{\Jason}[2][]{\noindentaftertodo}
\newcommand{\Steven}[2][]{\noindentaftertodo}
\newcommand{\Kitsing}[2][]{\noindentaftertodo}

\title{Structure-Aware Path Inference \\ for
Neural Finite State Transducers}

\author{
  Weiting Tan\quad Chu-Cheng Lin\thanks{Now at Google.}\quad Jason Eisner\\
  Department of Computer Science\\
  Johns Hopkins University \\
  \texttt{wtan12@jhu.edu \{kitsing, jason\}@cs.jhu.edu} \\[-12pt]
}

\begin{document}

\maketitle
\addtocounter{footnote}{-1}

\begin{abstract}
Neural finite-state transducers (NFSTs) form an expressive family of neurosymbolic sequence transduction models. An NFST models each string pair as having been generated by a latent path in a finite-state transducer. As they are deep generative models, both training and inference of NFSTs require inference networks that approximate posterior distributions over such latent variables.
In this paper, we focus on the resulting challenge of imputing the latent alignment path that explains a given pair of input and output strings (\eg during training).  We train three autoregressive approximate models for amortized inference of the path, which can then be used as proposal distributions for importance sampling.  All three models perform lookahead.  Our most sophisticated (and novel) model leverages the FST structure to consider the graph of future paths; unfortunately, we find that it loses out to the simpler approaches---except on an \emph{artificial} task that we concocted to confuse the simpler approaches.\looseness=-1
\end{abstract}

\begin{wrapfigure}{r}{0.35\linewidth}
    \centering
    \vspace{-13mm}
    \includegraphics[scale=0.45]{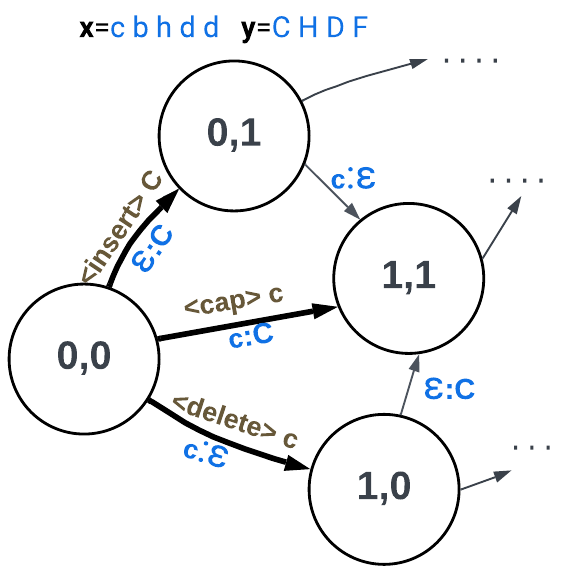}
    \caption{Marked finite-state transducer, all of whose paths generate input $\bm x$ and output $\bm y$. We show the transitions from the initial state, with their \textcolor{myblue}{input:output symbols} and \textcolor{mybrown}{marks}.  The same example appears in \cref{fig:method} without the marks shown.}
    \vspace{-1em}
    \label{fig::teaser}
\end{wrapfigure}

\section{Introduction}

Recent advances in applied deep learning have typically applied end-to-end training to homogeneous architectures such as recurrent neural nets (RNNs)~\cite{rnn}, convolutional neural networks \cite{convnet}, or Transformers \cite{transformer}. For small-data settings, however, end-to-end training can benefit from inductive bias through domain-specific constraints and featurization \cite{rastogi-etal-2016-weighting}.  In the case of sequence-to-sequence problems---\eg grapheme-to-phoneme \cite{knight-graehl-1998-machine} or speech-to-text \cite{mohri2008speech}---one technique is to use finite-state transducers (FSTs), which were widely used in NLP, speech recognition, and text processing before the deep learning revolution \cite{mohri-1997,EFSML-1999,allauzen2007openfst,mohri2008speech}. An FST's topology can be manually designed based on the task of interest.  In this paper, we design and compare inference networks for use with ``neuralized'' FSTs.

Neuralized FSTs (NFSTs)~\cite{lin-etal-2019-neural} abandon the Markov property to become more expressive than standard arc-weighted FSTs 
\cite{eisner-2002-parameter}.  A path's weight is  computed by some arbitrary neural model from the ordered string of marks encountered along the path. The marks on each arc provide features of the transduction operation carried out by that arc (as illustrated in \cref{fig::teaser}). 

However, the expressiveness of NFSTs comes at the cost of training efficiency. Modeling joint or conditional probabilities on observed string pairs $(\bm x, \bm y)$ requires imputing the latent NFST path $\bm z$ that aligns each observed input string $\bm x$ with its output string $\bm y$, as we review in \cref{sec::method}. Summing over these paths is expensive because NFSTs give up the Markov property that enables dynamic programming on traditional weighted FSTs \cite{mohri2002}. 
We must fall back on importance sampling. To this end, \Citet{lin-etal-2019-neural} built an autoregressive proposal distribution for these latent paths.\footnote{Better ensembles of weighted proposals can be jointly generated by using multinomial resampling (in particle filtering or particle smoothing), as \citet{lin-eisner-2018-neural} did in a simpler setting.  We do not pursue this extension here.\looseness=-1}  

Their proposal distribution---basically the SWS method described in \cref{sec::base_sampler} below---sampled a path through the NSFT from left to right.  At each step, the choice of the next arc was influenced by the prefix path sampled so far and by the suffixes of $\bm x$ and $\bm y$ that have yet to be aligned.  \textbf{In this paper, we extend this idea (\cref{sec::path_sampler}) to consider the graph of possible alignments of those suffixes (\cref{fig:method}), as determined by the NFST topology.}    
We evaluated the quality of the proposal distributions on three tasks:\looseness=-1
\begin{itemize}[noitemsep]
    \item (\texttt{tr}) reverse transliteration of Urdu words from the Roman alphabet to the Urdu alphabet
    \item (\texttt{scan}) compositional navigation commands paired with the corresponding action sequences
    \item (\texttt{cipher}) synthetic dataset created by enciphering the input text with certain patterns
\end{itemize}
Task examples are shown in \cref{table::dataset}, and more descriptions are available in \cref{app::dataset}. We compared our novel proposal distribution (\cref{sec::path_sampler}) to the approach of \cite{lin-etal-2019-neural} (\cref{sec::base_sampler}) and to an even simpler baseline (\cref{sec::attn_sampler}). Overall, it was difficult to get our novel method to work.  In the \texttt{tr} and \texttt{scan} tasks, it is apparently possible to choose the next arc well enough by the existing method of looking ahead to the \emph{unaligned} suffixes.  Perhaps the existing method learns to compare their lengths or their unordered bags of symbols.  We designed the \texttt{cipher} task to frustrate such heuristics, and there our novel method really was necessary, benefiting from its domain knowledge of possible alignments (the given FST).  But for the \texttt{tr} and \texttt{scan} tasks, our novel proposal distribution did considerably worse---perhaps our architecture was unnecessarily complicated and harder to train.  This raises questions about the necessity and wisdom of explicitly considering the graph of possible alignments for real-world tasks.

\section{Preliminaries: Neuralized Finite-State Transducers}\label{sec::method}

\subsection{Marked FSTs}
\begin{wrapfigure}{r}{0.35\linewidth}
    \centering
    \includegraphics[scale=0.4, trim=0 0 0 1cm]{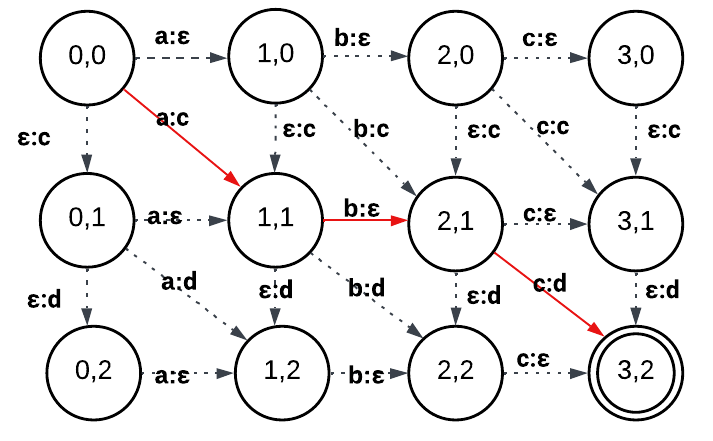}
    \caption{Directed graph constructed by composing an edit-distance MFST ${\cal T}$ with input $\bm x = abc$ and output $\bm y = cd$.  The marks are suppressed here, but see \cref{fig::teaser}.}
    \vspace{-1em}
    \label{fig::edit_distance}
\end{wrapfigure}
A marked FST, or MSFT, is a directed graph in which some states are designated as initial and/or final, and each arc is labeled with an input substring, an output substring, and a mark substring.  An \emph{generating path} in the MFST is any path $\bm z$ from an initial state to a final state.  It is said to generate the pair $(\bm x, \bm y)$ with mark string $\bm\omega$ if $\bm x, \bm y, \bm \omega$ respectively are the concatenations of the input, output, and mark substrings of $\bm z$'s arcs.\footnote{An ordinary FST omits the mark string, and the familiar FSA also omits the output string.}    

The mark string on a generating path of $\cal T$ provides domain-specific information about the path. It may record information about the states along the path, the symbols being generated (for example, their phonetic or orthographic properties), how the path aligns input and output symbols (that is, which symbols or properties are being edited), and the contexts of these aligned symbols (for example, whether they fall in the onset, nucleus, or coda of a linguistic syllable).  We may compose the MFST ${\cal T}$ with strings $\bm x, \bm y$ to obtain a restricted FST $\xTy$ whose generating paths correspond exactly to the paths in ${\cal T}$ that generate $(\bm x,\bm y)$, with the same marks.  A standard simple example is shown in \cref{fig::edit_distance}: input "abc" and output "cd" have been composed with a 1-state MFST whose arcs (which are self-loops) allow symbol insertions, deletions, and substitutions.  The generating paths are the paths from $(0,0)$ to $(3,2)$. The red path in \cref{fig::edit_distance} transforms "abc" to "cd" by deleting the second symbol of "abc" and substituting for the others. 

\subsection{Neuralized FSTs}\label{sec:nfst}

A neuralized finite-state transducer (NFST) is an MFST ${\cal T}$ paired with some parametric scoring function $\scorer$ that maps any generating path's mark string to a non-negative weight. Given an appropriate ${\cal T}$, this defines an unnormalized probability distribution $\scorer$ over the paths. See \cref{app::NFST_def} for a formal definition.  

In practice, we will estimate $\scorer$ by estimating its parameters $\theta$.
The unnormalized probability of a string pair is obtained by summing over the paths $\bm z$ that might have generated that string:
\begin{equation}
    \tilde{p}_\theta({\bm x},{\bm y}) \defeq \textstyle\sum_{{\bm z} \in \xTy}\scorer(\bm z)
\label{eq:nfst-rel-def}
\end{equation}
Given the pair $(\bm x, \bm y)$, the posterior distribution over the latent generating path $\bm z \in \xTy$ is $p_\theta(\bm z \mid \bm x, \bm y) \defeq \tilde{p}_{\theta}({\bm z})/\tilde{p}_{\theta}({\bm x}, {\bm y})$. We emphasize that $\tilde{p}_\theta(\bm z)$ depends on $\bm z$ only through its mark string. 

Computing the quantity $\log \tilde{p}_{\theta}({\bm x}, {\bm y})$ is crucial for training $\theta$. However, \cref{eq:nfst-rel-def}'s 
marginalization over mark strings $\bm z$ is in general intractable. 
We resort to using a Monte Carlo variational lower bound, which imputes $\bm z \sim p_\theta(\cdot \mid \bm x, \bm y)$ by importance sampling using a neural proposal distribution $q_\phi(\bm z \mid \bm x, \bm y) \approx p_\theta(\bm z \mid \bm x, \bm y)$.  In \cref{app::NFST_train} we describe a procedure for jointly training $p_\theta$ and $q_\phi$, making use of the importance weighting estimator \cite{iwae} and making certain assumptions about $\cal T$ and $p_\theta$.\looseness=-1

Our focus in this paper is to consider different parametric forms for the distribution $q_\phi$ over paths that generate $(\bm x, \bm y)$.  To simplify our study, we assume that $\theta$ is given, and only train $\phi$ to minimize the divergence 
$\text{KL}(p_\theta\mathrel{\Vert}q_{\phi}) = \underset{\bm z\sim p(\cdot \mid \bm x, \bm y)}{\mathbb{E}}[-\log q_{\phi}(\bm z \mid \bm x, \bm y)]$ by following its gradient (or rather, the biased estimate of its gradient that we obtain by normalized importance sampling, sample size 16).

\section{Three Proposal Distributions}\label{sec::sample}
\begin{figure*}[t]
    \centering
    \begin{tabular}{c}
        \includegraphics[scale=0.51, trim=6mm 0mm 0mm 8mm]{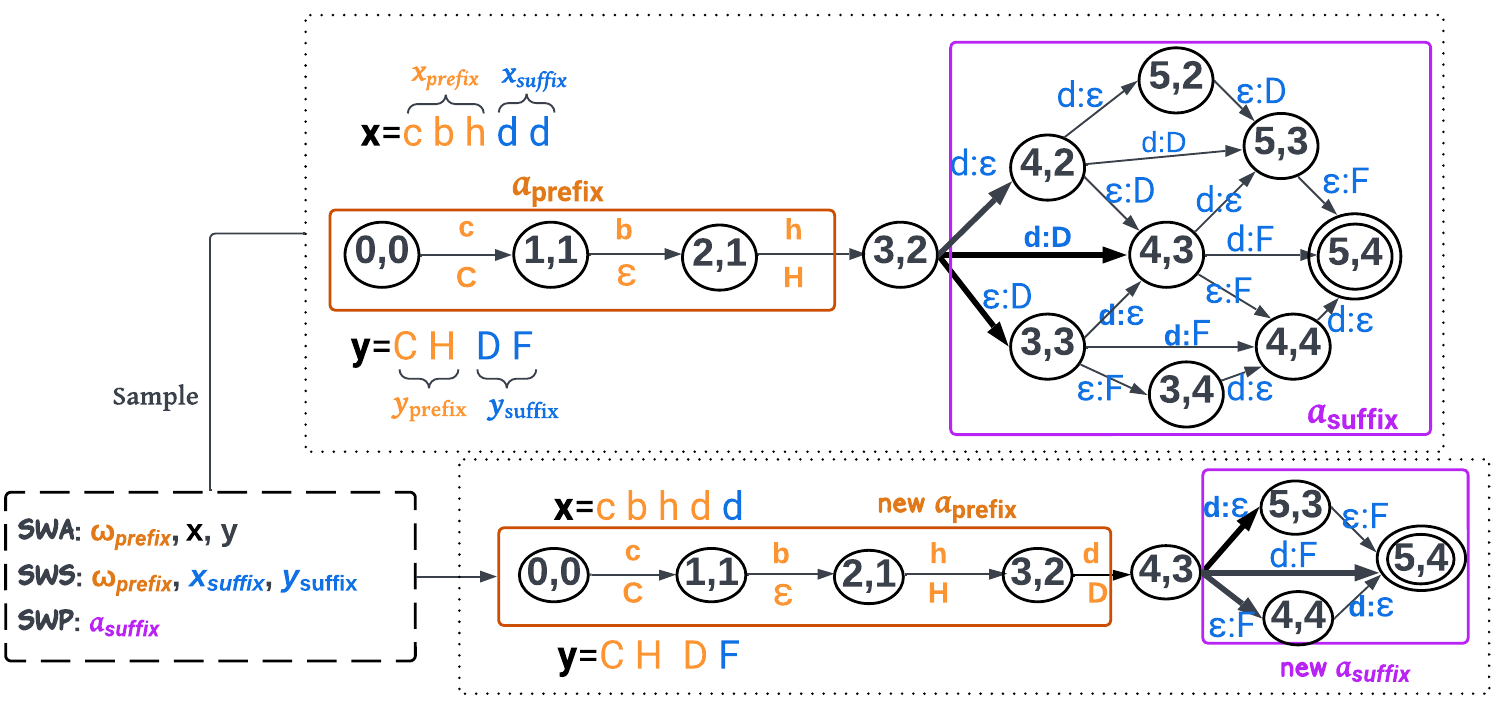} \\
    \end{tabular}
    \caption{
    Choosing a generating path $\bm a$ in $\Txy$ with mark string $\bm \omega$.  (Note that $\bm a$ ranges over paths in $\Txy$, whereas $\bm z$ ranges over paths in $\cal T$. Marks are not shown.)  The top graph shows that after choosing the first 3 arcs \textcolor{prefixcolor}{$\bm a\prefix=a_1 a_2 a_3$}, reaching state $(3,2)$, the sampler must choose a suffix path through the subgraph \textcolor{asuffix}{$\bm a\suffix$} of $\Txy$ that is reachable from $(3,2)$.  Its choices for $a_4$ are the three thick out-arcs from $(3,2)$; choosing the arc to $(4,3)$ yields the reduced graph of possible paths $\bm a$ at the bottom.  To make its stochastic choice, the sampler (SWA, SWS, or SWP) conditions on certain properties of the top graph, shown in the dashed box at the left.  Here \textcolor{prefixcolor}{$\bm x\prefix,\bm y\prefix,\bm\omega\prefix$} refer to the labels on \textcolor{prefixcolor}{$\bm a\prefix$}, while \textcolor{xysuffix}{$\bm x\suffix,\bm y\suffix$} are the remaining portions of $\bm x,\bm y$.  SWP's choice is determined by a global backward pass on $\Txy$ in which the probabilities of the out-arcs depend only on \textcolor{asuffix}{$\bm a\suffix$}.
    }\label{fig:method}
   \vspace{-3mm}
\end{figure*}

We explore three distribution families $q_\phi(\bm z \mid \bm x, \bm y)$, sketched in \cref{fig:method}.  Like $p_\theta$ itself, each $q_\phi$ family is insensitive to the specific topology and labeling of $\cal T$.  Any MFST that was \emph{equivalent} to $\cal T$ in the sense of generating the same set of $(\bm x, \bm y, \bm \omega)$ triples---that is, the same regular 3-way relation---would give the same parametric proposal distribution $q_\phi$.  Sampling from $q_\phi$ in each case is done by sampling a mark string $\bm\omega$ and using it to identify a path $\bm z$ in $\cal T$.  To make this identification possible, we henceforth assume that distinct paths in $\cal T$ with the same $\bm x, \bm y$ always have distinct mark strings.  (A stronger assumption is already needed for the particular model $\tilde{p}_\theta$ that we spell out in \cref{app::NFST_train}.)  

Let $\Txy$ be a version of $\xTy$ that has been determinized with respect to the mark tape and then minimized.\footnote{Brief technical details 
\cite[see e.g.][]{reutenauer-1990,mohri-1997,allauzen2007openfst}: 
Treat $\cal T$ as a finite-state automaton over the mark alphabet, weighted by (input, output) string pairs.  $\xTy$ can be determinized (made subsequential) because it is unambiguous (due to our assumption above) and has bounded variation (since all mark strings map to the fixed strings $\bm x$ and $\bm y$) or equivalently has the twins property (since for the same reason, all cycles must produce empty input and output).  We remark that only the SWS sampler really requires the (input, output) weights, as they guide its sampling of a mark string.  The SWA and SWP samplers could simply drop them from $\xTy$ and apply ordinary unweighted determinization and minimization \cite[see][]{hopcroft-ullman-1979}.}
Then $\Txy$ is a canonical MFST that expresses the same relation as $\xTy$, while guaranteeing that (a) each arc's mark substring has length 1, (b) the out-arcs from a state are labeled with different marks, and (c) every choice of out-arc can lead to a final state.  

It follows that we can sample a mark string $\bm \omega$ by autoregressively sampling a generating path $\bm a$ in $\Txy$, at each step requiring only a distribution over the out-arcs from the current state, or equivalently, over their distinct marks.  We abuse notation and write $q_\phi$ for $q_\phi(\bm a \mid \Txy)$, $q_\phi(\bm \omega \mid \Txy)$, and $q_\phi(\bm z \mid \bm x, \bm y)$.

Since we use the standard minimization construction, $\Txy$ is \emph{reduced}, meaning that the input and output symbols on a path appear ``as soon as possible'' \cite{reutenauer-1990}.  This standardizes how the symbols are distributed along the path, ensuring that $\bm x\suffix, \bm y\suffix$ are well-defined in the SWS sampler (\cref{sec::base_sampler}).

\subsection{Sampler with Attention (SWA)}\label{sec::attn_sampler}
The SWA sampler proposes the next mark $\omega_t$ in $\bm\omega$ by considering (1) the marks $\bm\omega_{<t} = \omega_1\cdots\omega_{t-1}$ already chosen and (2) an assessment of which marks might be chosen in future. For (1), we use a recurrent neural network to encode $\bm\omega_{<t}$ into a vector $\vh_{t-1}$. 
For (2), an attention mechanism is employed to mine information from the raw strings $(\bm x, \bm y)$.    
More formally, SWA samples from
\begin{align}\label{eq:swa}
        q_\phi(\omega_t \mid \bm \omega_{<t}, \Txy) 
        &\propto \exp (W\,[1;\vh_{t-1}; \text{Att}(\vh_{t-1}, \text{enc}(\bm x,\bm y))])
\end{align}
meaning a softmax distribution over just the marks $\omega_t$ that $\Txy$ allows to follow $\bm \omega_{<t}$.
\Cref{eq:swa} applies the learned matrix $W$ to the concatenation of the state $\vh_{t-1}$, which encodes $\bm\omega\prefix$, and $\text{Att}(\cdots)$ which uses $\vh_{t-1}$ to attend to $(\bm x, \bm y)$ (not necessarily just $\bm x\suffix, \bm y\suffix$ in \cref{fig:method}
We compute $\vh_{t-1} \in \mathbb{R}^d$ with a left-to-right GRU \cite{gru}%
, as in \citet{lin-eisner-2018-neural}.
We compute $\text{enc}(\bm x,\bm y) \in \mathbb{R}^{n\times d}$ by applying a separate bidirectional GRU%
\footnote{Other representation methods like Transformer encoder could also be used.} to the concatenation $\bm x\# \bm y^R$, whose length we denote by $n$.\footnote{Inspired by \cite{sutskever2014sequence}, we use the reversed output string $\bm y^R$ so that the end of $\bm x$ and the end of $\bm y$ are adjacent.  
This setup makes it potentially easier for $\text{enc}(\bm x, \bm y)$ to consider alignments between $\bm x\suffix$ and $\bm y\suffix$.} 
We then use $\vh_{t-1}$ as a softmax-attention query over the $n$ encoded tokens of $\bm x\# \bm y^R$:
\begin{equation}\label{eq:att}
    \text{Att}(\vh_{t-1}, \text{enc}(\bm x,\bm y)) = \sum_{i=1}^{n} a_i\,\text{enc}_{i}(\bm x,\bm y), \text{\ \ \ \ where } a_i = \frac{\exp(\vh_{t-1}\cdot\text{enc}_{i}(\bm x,\bm y))}{\sum_{j=1}^{n} \exp (\vh_{t-1}\cdot\text{enc}_{j}(\bm x,\bm y))}
\end{equation}

\subsection{Sampler with State Tracking (SWS)}\label{sec::base_sampler}

The SWS sampler is simpler than SWA (no attention), but it takes care to consider only the suffixes of $\bm x$ and $\bm y$ that remain to be aligned.  The prefix path of $\Txy$ marked with $\bm \omega_{<t}$ has generated aligned input and output strings $\bm x_{< t}, \bm y_{< t}$ (which may not have length $t$).  Let $\bm x_{\geq t}$ and $\bm y_{\geq t}$ denote the unaligned suffixes of $\bm x, \bm y$, and encode them into vectors by $\text{enc}_\text{x}$ and $\text{enc}_\text{y}$, which are separate right-to-left GRUs. SWS samples from
\begin{align} 
        q_\phi(\omega_t \mid \bm \omega_{<t}, \Txy) 
        &\propto \exp(W\,[1;\vh_{t-1}; \text{enc}_\text{x}(\bm x_{\geq t}) ; \text{enc}_\text{y}(\bm y_{\geq t})])
\end{align}
As before, this is a softmax over choices of $\omega_t$ that are legal under $\Txy$; other marks have probability 0.\looseness=-1

The SWS method is directly inspired by \cite{lin-etal-2019-neural}. It is a relatively weak model, as the three information sources $\bm\omega_{<t}$, $\bm x_{\geq t}$, and $\bm y_{\geq t}$ contribute \emph{independent} summands to the logits of the possible marks.  There is no additional feed-forward layer that allows these sources to interact.  In contrast, SWA mixed all three sources by having an encoding of $\bm\omega_{<t}$ attend to a joint encoding of $\bm x\# \bm y^R$.

\subsection{Sampler with Path Structures (SWP)}\label{sec::path_sampler}

The main contribution of this work is the SWP sampler. 
It assigns embeddings and weights to the arcs of $\Txy$ and uses this to define a distribution over its paths $\bm a$, yielding a distribution over mark strings $\bm \omega$.  So it samples $\bm a \in \Txy$ and then returns the
$(\bm x,\bm y)$-generating path $\bm z \in \cal T$ with the same marks.

SWP's proposal distribution assigns weights to the arcs of $\Txy$ and \textbf{treats it as a weighted FST (WFST)}.  Recall that in the NFST $\cal T$, paths are scored globally using $\tilde{p}_\theta(\bm z)$.  A WFST is simpler: each arc $s \to s'$ has a fixed weight $w_{s \to s'}$, and the weight of a path $\bm a$ is the product of its arc weights.\footnote{In general there could be multiple $s \to s'$ arcs, but for simplicity, we gloss over this in the notation.}
This makes exact path sampling tractable: if the path 
$\bm a_{<t} = a_1 \cdots a_{t-1} = \bm $ ends at state $s$, then the autoregressive probability of choosing $a_t$ to be the arc $s \to s'$ depends only on $s$ (a Markov property): 
\begin{equation}\label{eq:swp_sample}
     q_\phi(s \to s' \mid \bm a_{<t}, \Txy)
     = q_\phi( s \to s' \mid s)
     = w_{s \to s'}\beta(s') \,/\, \beta(s)
\end{equation}
where $\beta(s)$ denotes the ``backward weight''---the total weight of all paths from $s$ to a final state.  

The $\beta(s)$ values are the solution to the  system of linear equations $\beta(s) = (\sum_{s'} w_{s\to s'}\beta(s')) + \mathbb{I}(s\text{ is final})$.
If $\Txy$ is acyclic, this is an acyclic recurrence that can be solved in linear time by the backward algorithm \cite{baum-welch-1970,rabiner-1989}.  

But where do the WFST arc weights $w_{s\to s'}$ come from?  We will assume $\Txy$ is acyclic and use a similar recurrence to define arc embeddings $\vec{e}_{s\to s'}$ and state embeddings $\vec{e}_s$ that yield the arc weights $w_{s\to s'}$.  \Cref{alg::beta} computes these along with $\beta(s)$ and $q_\phi(s \to s' \mid s, \Txy)$, all in a single pass.  

The state embedding $\vec{e}_s$ is a LatticeRNN-like embedding \cite{sperber-etal-2017-neural} that attempts to summarize the mark strings of all suffix paths from $s$.  A graph of such paths is shown as $\bm a\suffix$ in \cref{fig:method}. Suffix paths with higher WFST weight will have more influence on the summary, thanks to the weighted average at \cref{line:stateemb}.  Similarly, the arc embedding $\vec{e}_{s\to s'}$ attempts to summarize all suffix paths of the form $s \to s' \to \cdots$.  Put another way, it encodes the mark on $s\to s'$ in a way that considers its possible right contexts.\looseness=-1

.

\vspace{-5mm}
\begin{algorithm}
\caption{Constructing WFST arc weights $w$, backward weights $\beta$, and transition probabilities $q$}\label{algo:embed-update}
\begin{algorithmic}[1] 
\Statex Input: An acyclic MFST $\Txy$, 
\For{$s \leftarrow \text{states}(\Txy)$ in reverse topological order}
\State $\beta(s) = \mathbb{I}(s\text{ is final})$
\For{each out-arc $s \rightarrow s'$ (with mark $\omega$)}  \Comment{note that $s'$ has already been visited}
\State $\vec{e}_{s\to s'} = \sigma(U\, [1; \vec{e}_\omega; \vec{e}_{s'}])$ \Comment{$U$ and mark embeddings $\vec{e}_\omega$ are learned (part of $\phi$)}\label{line:embarc}
\State $w_{s\to s'} =  \exp(\vec{w} \cdot \vec{e}_{s\to s'})$ \Comment{arc weight; $\bm w$ is learned} \label{line:arcweight}
\State $\beta(s) \pluseq w_{s \rightarrow s'} \beta(s')$ \Comment{backward algorithm}
\EndFor
\State $q_\phi(s \rightarrow s' \mid s) = 
w_{s \to s'}\beta(s') / \beta(s)$ \Comment{transition probability (\cref{eq:swp_sample})}
\State $\vec{e}_s = \sum_{s'} q_\phi(s \rightarrow s' \mid s)\,\vec{e}_{s \rightarrow s'}$ \Comment{state embedding is weighted average of arc embeddings}\label{line:stateemb}
\EndFor
\State \Return all transition probabilities $q_\phi(s \rightarrow s' \mid s)$
\end{algorithmic}
\label{alg::beta}
\end{algorithm}
\vspace{-3mm}

After running \cref{alg::beta}, we can use $q_\phi$ to sample a path $\bm a$ of $\Txy$ from left to right, yielding its mark string $\bm \omega$.  Choosing the next arc $a_t$ is tantamount to choosing its mark as the next mark $\omega_t$.

This formulation of SWP does have some limitations relative to SWA and SWS.  The right-to-left embedding update at \cref{line:embarc} is RNN-style; one would have to add a GRU-style forget gate to make it comparable.  Another limitation of SWP is that its choice of $\omega_t$ is \emph{not} influenced by the embedding $\vec{h}_{t-1}$ of $\bm\omega_{t-1}$.  An obvious fix would be to replace $\vec{w}$ with $\vec{h}_{t-1}$ at \cref{line:arcweight}.  However, then sampling at each step $t$ would require re-running \cref{alg::beta} with the new $\vec{h}_{t-1}$ to re-embed and re-weight the sub-MFST that is reachable from the current state $s$ (shown in \cref{fig:method}).  A cheaper variant would run \cref{alg::beta} only at the start and never change the embeddings or weights, but have step $t$ choose arc $s\to s'$ with probability proportional to $\exp(\vec{h}_{t-1}\cdot\vec{e}_{s\to s'})\beta(s')$ or perhaps just $\exp(\vec{h}_{t-1}\cdot\vec{e}_{s\to s'})$.  

\section{Training and Evaluation of Proposal Samplers}\label{sec::eval}

Recall that our $q_\phi$ chooses $\bm z$ in $\cal T$ (by sampling $\bm \omega$ from $\Txy$).
To evaluate whether $q_\phi(\bm z \mid \bm x, \bm y) \approx p_\theta(\bm z \mid \bm x, \bm y)$ as desired,
we follow \citet{lin-eisner-2018-neural} and consider the \emph{exclusive} KL divergence
\begin{equation}\label{eq:KLexc}
    \text{KL}(q_{\phi}\mathrel{\Vert}\scorer) = \mathbb{E}_{\bm z \sim q_{\phi}(\cdot \mid \bm x, \bm y)}[\log q_\phi(\bm z \mid \bm x, \bm y) - \log p_\theta(\bm z \mid \bm x, \bm y)]
\end{equation}
However, since the normalization term \labelcref{eq:nfst-rel-def} is hard to compute, we drop it and compute the resulting \textit{Partial KL}, averaging it over a held-out test dataset $\mathcal{D}$ (and estimating the expectation by sampling):
\begin{equation}\label{eq:partial-kl}
    \frac{1}{|\mathcal{D}|} \textstyle\sum_{(\bm x_i,  \bm y_i)\in \mathcal{D}} \mathbb{E}_{\bm z \sim q_{\phi}(\bm z \mid \bm x_i, \bm y_i)}\left[\log q_{\phi}(\bm z \mid \bm x_i, \bm y_i) - \log \scorer(\bm z)\right]   
\end{equation}
To make the comparison fair, we need to enforce that the dropped normalization term \labelcref{eq:nfst-rel-def} is the same for different samplers. Thus, we evaluate our different samplers with the same \emph{frozen} model $\scorer$, only training the sampler parameters $\phi$.  We train $\phi$ to minimize the \emph{inclusive} KL divergence, as mentioned at the end of \cref{sec:nfst}, to avoid the instability of minimizing the exclusive divergence \labelcref{eq:partial-kl} by the high-variance \textsc{reinforce} estimator.  The reason that we still evaluate with \emph{exclusive} KL is that it serves our actual goals for $q_\phi$: if we were training $\theta$ to minimize the variational log-likelihood \labelcref{eq::loss} in \cref{app::NFST_train}, then a $q_\phi$ with smaller exclusive KL would obtain a tighter variational bound.

We obtained our frozen $\scorer$ by alternately optimizing $\theta$ (via \labelcref{eq::loss}) and the $\phi$ of an SWA sampler $q_\phi$ (again trained via inclusive KL, for stability, even though exclusive KL would give a tighter bound \labelcref{eq::loss}), simplifying the latter to replace GRUs with RNNs.
We then discarded this sampler. (One might expect this choice of $\scorer$ to benefit the SWA sampler, but as we will see below, 
it never performed best.)

\section{Experiments}\label{sec::experiment}
We conducted our experiments on three datasets (described in \cref{app::dataset}): \texttt{tr}, \texttt{scan}, and \texttt{cipher}.
\Cref{table::result} has the main results. \Cref{table::sample_length} shows the importance sampling estimate of the expected mark string length $\mathbb{E}_{(\bm x,\bm y)\sim \mathcal{D}}\left[\mathbb{E}_{\bm z \sim p_\theta(\cdot \mid \bm x, \bm y)}\left[\, |\bm z_\Omega| \,\right]\right]$
and \cref{table::ess} shows the (de-duplicated) effective sample size.
For \texttt{tr} and \texttt{scan} tasks, the mark string length is fixed regardless of alignment, as the FST design only allows deletion/insertion. For \texttt{cipher}, however, SWP is able to discover higher-probability alignments with more substitutions and shorter mark strings, both by identifying the correct cipher and by aligning the strings more accurately under that cipher.  This improves both \textit{Partial KL} and estimated length. Qualitative examples of proposals from different samplers are presented in \cref{app::qual}.

\begin{figure}[ht]
    \vspace{-0.3em}
    \begin{subtable}{0.49\linewidth}
        \centering
        \begin{tabular}{ccccc}
            \toprule
            Sampler & TR & Scan & Cipher & Avg \\
            \midrule
            SWA & 23.3 & 38.1 & 101.3   & 54.3 \\
            SWS & \textbf{21.8} & \textbf{25.4} & 87.1 &   \textbf{44.7} \\
            SWP & 38.1 & 102.9 &  \textbf{73.9} &  71.7 \\
            \bottomrule
        \end{tabular}
        \caption{\Cref{table::result}. \textit{Partial KL} divergence comparison (lower values are better) for different proposal samplers.}
        \label{table::result}
    \end{subtable}
    \hspace{0.2cm}
    \begin{subtable}{0.49\linewidth}
        \centering
        \begin{tabular}{llll}
            \toprule
            Sampler & TR & Scan &  Cipher\\
            \midrule
            SWA & 24.3 & 65.6 & 68.4 \\
            SWS & 24.3 & 65.6 & 67.6 \\
            SWP & 24.3 & 65.6 & 59.9 \\
            \bottomrule
        \end{tabular}
        \caption{\Cref{table::sample_length}. Expected  mark string length under $p_\theta$, estimated by importance sampling with proposals $\sim q_\phi$.}
        \label{table::sample_length}
    \end{subtable}%
\end{figure}

For \texttt{tr}, the true alignment between Urdu and English scripts is monotonic, and the training/test sets exhibit similar length distributions. Here the SWA sampler performed comparably to the SWS sampler, suggesting that a basic attention mechanism suffices on this task.
SWP showed no advantage in this case (indeed, it did worse), probably because simple symbol-counting and local-lookahead heuristics are effective at guessing the next step, and these can be captured by both SWA and SWS architectures.

In the \texttt{scan} dataset, where the true alignment is non-monotonic and test examples are longer than training examples, the SWS sampler outperforms the SWA sampler. Presumably SWS is able to find better alignments---that is, generating paths with higher scores $\tilde{p}(\bm z)$---by looking ahead into the near future of the unaligned suffixes $\bm x_{\geq t}, \bm y_{\geq t}$ to see how to complete the partial alignment summarized by $\vec{h}_{t-1}$.  
We had expected that SWP might do even better by globally aligning $\bm x, \bm y$ using a WFST, but evidently it completely failed to learn the alignment (see \cref{tab:scan_examples}), perhaps due to uninformed initialization\footnote{It might have helped to do \emph{supervised} pre-training of $q_\phi$ on $(\bm x, \bm y, \bm \omega)$ triples drawn unconstrained from $\tilde{p}_\theta$.} or the WFST's Markov property that ignores $\bm z_{<t}$.  Indeed, SWP did even worse than a baseline ``no-lookahead'' model (see \cref{table::no_lookahead}) we trained to predict the next mark $\omega_t$ from only $\vec{h}_{t-1}$.

Moving to the \texttt{cipher} task, however, the SWP sampler excelled.  The FST's topology is more informed in this case, and plays a crucial role in considering possible alignments of $\bm x_{\geq t}, \bm y_{\geq t}$ (especially at $t=1$, to identify which cipher should be used). 
The other samplers struggle here without this, as simple heuristics no longer work.  
A hyperparameter sweep for this dataset
(see \cref{fig:ablation_dim}) reinforces the importance of alignment information: even the smallest model for the SWP sampler (hidden dimension=64) outperforms SWS and SWA samplers based on much larger models.

\section{Conclusion}\label{sec::conclusion}

In this paper, we explored proposal distributions over paths in a graph. 
We found that while the proposal distribution could sometimes benefit from examining the structure of the graph, our proposal distribution that did so proved hard to train.  As a result, except on a particularly challenging artificial task, our structure-aware sampler performed worse than its less complex, non-structure-aware alternatives. 
One might try improved training methods to minimize either inclusive KL, such as DPG\textsubscript{off} \cite{parshakova-2019}, or exclusive KL, such as PPO with entropy bonus \cite{schulman2017proximal}.
It is also possible that variant architectures would succeed better, as discussed in \cref{sec::path_sampler}.

\bibliography{main}
\bibliographystyle{apalike}

\newpage

\appendix
\section*{\LARGE{Supplementary Material}}

\begin{table}[ht]
    \centering
    \footnotesize
    \begin{tabular}{cl}
    \textbf{Appendix Sections}    & \textbf{Contents}  \\ \toprule
    \cref{app::dataset}     &  \begin{tabular}[c]{@{}l@{}} Description of datasets\end{tabular} \\ \midrule
    \cref{app::NFST_def}     &  \begin{tabular}[c]{@{}l@{}} Definition of Neuralized Finite State Transducers (NFSTs)\end{tabular} \\ \midrule
    \cref{app::NFST_train}     &  \begin{tabular}[c]{@{}l@{}} Parametrizing and training NFST models\end{tabular} \\ \midrule
    \cref{app::hyperparams}     &  \begin{tabular}[c]{@{}l@{}} Hyperparameter settings \end{tabular}\\ \midrule
     \cref{app::qual}     &  \begin{tabular}[c]{@{}l@{}}Qualitative comparison of samplers\end{tabular} \\ \midrule
     \cref{app::no_lookahead}     &  \begin{tabular}[c]{@{}l@{}}Comparison with no-lookahead baseline sampler\end{tabular} \\ \midrule
     \cref{app::limitation}     &  \begin{tabular}[c]{@{}l@{}}Discussion of limitations \end{tabular} \\ \bottomrule
\end{tabular}    
\end{table}

\section{Datasets}\label{app::dataset}
In this section, we describe the datasets in more detail. Examples of the 3 tasks are shown in \cref{table::dataset} and statistics in \cref{table::data_stats}. 

\begin{table*}[ht]
\centering
\small
\begin{tabular}{p{0.07\textwidth}p{0.42\textwidth}p{0.42\textwidth}}
\hline
\textbf{Dataset} & \textbf{Input}& \textbf{Output}\\
\hline
\texttt{tr} & a s a m a r & (some urdu text) \\
\hline
\texttt{scan} &look right twice after jump left & I\_TURN\_LEFT I\_JUMP I\_TURN\_RIGHT I\_LOOK I\_TURN\_RIGHT I\_LOOK \\
\hline
\texttt{cipher} & g c c i n f j n c g & i i r d i t y s \\
\hline
\end{tabular}
\caption{Examples of datasets used.}
\label{table::dataset}
\end{table*}

\begin{figure}[h]
    \includegraphics[scale=0.79, trim=-8em 0.3em 0em 0em]{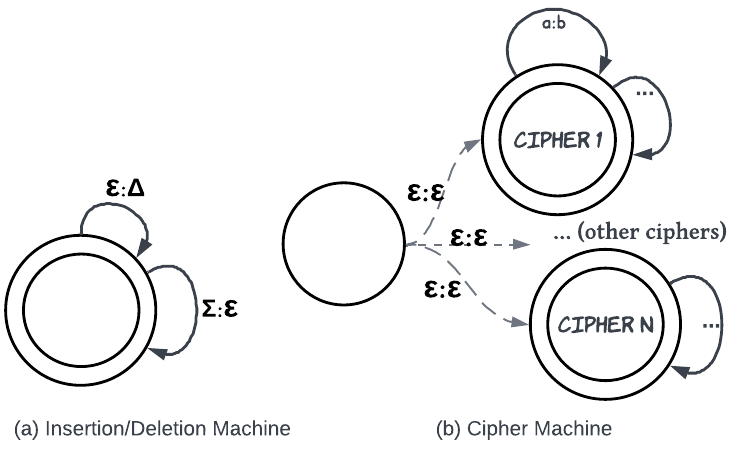}
    \caption{MFST topologies $\cal T$ used for different tasks. Here $\Sigma$ and $\Delta$ are the alphabets of possible input and output symbols.  An arc with input, output, and mark substrings $\bm x, \bm y, \bm \omega$ would ideally be displayed with the label $\bm x:\bm y / \bm \omega$, but in this figure we suppress the marks $\bm \omega$. \\ 
    (a)~\texttt{tr} and \texttt{scan} only use this deletion/insertion topology, which freely generates input symbols $a \in \Sigma$ (via deletion arcs $a:\varepsilon$) interleaved arbitrarily with output symbols $b \in \Delta$ (via insertion arcs $\varepsilon:b$).  Thus, it generates all $(\bm x,\bm y)$ pairs with all possible alignments.   \\(b)~\texttt{cipher} begins with the random-cipher MFST shown in the drawing.  That MFST deterministically replaces each input symbol $a \in \Sigma$ with a specific other symbol $b \in \Delta$ (via a substitution arc $a:b$), according to a cipher selected nondeterministically from $N=5$ possible ciphers.  The \texttt{cipher} task composes this with a deletion/insertion/copy MFST, which has a single state (just as in the \texttt{tr} task) and self-loop arcs of the form $b:\varepsilon$ (deletion), $\varepsilon:b$ (insertion), and $b:b$ (copy) for all $b \in \Delta$. Thus, some of the enciphered input symbols are deleted from the output, while some are copied and some other output symbols are inserted.  We implement the composition of two MFSTs so that when $a:b$ with marks $\bm\omega_1$ from the cipher MFST composes with $b:\varepsilon$ or $b:b$ with marks $\bm\omega_2$ from the deletion/insertion/copy MFST, the resulting arc (whose input:output is $a:\varepsilon$ or $a:b$ respectively) will have marks $\bm\omega_1\bm\omega_2$, in that order.}
    \label{fig::fsm}
\end{figure}

\begin{table}[!ht]
\centering
\begin{tabular}{lcccccc}
\toprule
Task & Split & Input Length & Output Length & Avg \#States & Avg \#Arcs  \\ 
\midrule
tr, vocab size=177 & Train & 7.2 & 6.9 & 232 & 303  \\ 
& Valid & 7.3 & 7.0 & 237 & 310  \\ 
& Test & 3.0 & 6.8 & 220 & 287  \\ \midrule
scan, vocab size=35 & Train & 8.1 & 12.0 & 285 & 366  \\ 
& Valid & 8.0 & 11.5 & 271 & 349  \\
& Test & 9.0 & 25.3 & 658 & 854  \\ \midrule
cipher, vocab size=67 & Train & 6.2 & 6.4 & 1028 & 1191  \\ 
& Valid & 6.1 & 6.3 & 992 & 1149  \\
& Test & 12.0 & 12.2 & 3989 & 4645 \\ \bottomrule
\end{tabular}
\caption{Data statistics for different tasks and splits. The vocab size refers to the size of the mark alphabet. The state and arc counts refer to $\Txy$ before minimization, averaged over $(\bm x, \bm y)$ pairs in the dataset.  Note that a single edit operation such as ``delete $a$'' may be carried out in $\cal T$ by a single arc with input $a$, output $\varepsilon$, and mark string ``<delete><a>'', but each instance of this operation in $\Txy$ will require two arcs with an intermediate state, since each arc of $\Txy$ has exactly one mark.  Before minimization, such intermediate states have only one out-arc, which reduces the arc-to-state ratio.%
}
\label{table::data_stats}
\end{table}

\paragraph{Transliteration (\texttt{tr})}
For the reverse transliteration task, we use the English-Urdu section from the Dakshina \cite{roark-etal-2020-processing} dataset. The topology is shown in \cref{fig::fsm}a, which is a deletion/insertion MFST. Such an MFST allows for any arbitrary transliteration from the Roman alphabet $\Sigma$ to the Urdu alphabet $\Delta$. Examples are shown in \cref{table::dataset,tab:tr_examples}.

We hope that after training, the model prefers---and the sampler proposes---paths where corresponding English and Urdu tokens are deleted and inserted close to each other.
Note that our MFST $\cal T$ does not include substitution arcs like $a:bc$.  It would accomplish this substitution by a sequence such as $a:\varepsilon, \varepsilon:b, \varepsilon:c$, and $\tilde{p}_\theta$ can be trained to favor such sequences.  This saves us from the need to enumerate all reasonable substitutions within $\cal T$.  We do not even include copy arcs like $a:a$, since the Urdu and English alphabets are different.    

\paragraph{Navigation Commands (\texttt{scan})}
We apply the same deletion/insertion topology to the \texttt{scan} dataset. The dataset contains command instructions in an English-like artificial language as input and action sequences as output. Alignments are nonmonotonic.  Examples are shown in \cref{table::dataset,tab:scan_examples}.

In this dataset, test samples are longer on average than training samples, requiring the model $\tilde{p}_\theta$ to actually learn the correspondence between commands and actions to generalize well to test data.  Whether or not $\tilde{p}_\theta$ succeeds in generalization, our goal in this paper is to make $q_\phi$ approximate the conditionalization of $\tilde{p}_\theta$ on test data.

\paragraph{Cipher (\texttt{cipher})}
We synthetically create a cipher dataset by composing a cipher MFST with an deletion/insertion/copy MFST (see \cref{fig::fsm}). Each of the 5 ciphers is chosen randomly (through random permutation using the Fisher-Yates shuffle algorithm \cite{fisher_yates1938}). As shown in \cref{fig::fsm}b, the initial state has $\varepsilon$ arcs to 5 cipher states; these arcs are marked with distinct marks.  The $i$\textsuperscript{th} cipher state has 26 arcs corresponding to the enciphering process $x \rightarrow \sigma_i(x)$ for an input alphabet of size 26, and the $i$\textsuperscript{th} cipher function $\sigma_i(\cdot)$ is chosen randomly.  As explained in the caption to \cref{fig::fsm}, we then compose the cipher MFST with a deletion/insertion/copy MSFT that introduces noise into the ciphertext.  The point of this construction is that a cipher-dependent alignment is now required.
Similar to \texttt{scan}, we intentionally make the training samples shorter on average compared to the test samples, so that $\tilde{p}_\theta$ must learn to generalize.

Example mark strings are shown in \cref{tab:cipher_examples}.
Note that the sampler $q_\phi$ must begin by predicting a mark that indicates which of the 5 ciphers will be used.  All of our sampler designs use lookahead to make this prediction, but SWP has an advantage, because it can examine the entire graph $\bm a\suffix$ of possible $\bm x$-to-$\bm y$ alignments under all 5 ciphers.  This graph is constructed with explicit knowledge of the FST topology, which the other samplers lack.

\section{Neuralized Finite-State Transducers}\label{app::NFST_def}
A neuralized finite state transducer (NFST) \cite{lin-etal-2019-neural} is defined as a pair $({\cal T}, \scorer)$ where ${\cal T}$ is a marked finite-state transducer (MFST) and $\scorer$ is a mark string scoring function parmeterized by $\theta$. 

An MFST can be regarded as having three left-to-right tapes, one for input symbols, one for output symbols, and one for the mark symbols. More formally, 
$${\cal T}=(\Sigma, \Delta, \Omega, Q, E, q_{\text{init}}, F)$$ where $\Sigma, \Delta,  \Omega$ are alphabets of input symbols, output symbols, and mark symbols. $Q$ is a finite set of states and $E\subseteq Q \times (\Sigma^* \times \Delta^* \times \Omega^*) \times Q$ is a finite multiset of arcs. $q_{\text{init}}$ is the initial state and $F$ is a set of final states. Note that the marks on an arc may mention the input and output on that arc, but they can also mention other features that are consumed by $\tilde{p}_\theta$; for example, the symbol "<C>" can be used to mark every arc that outputs a consonant.  A generating path $\bm z$ in $\cal T$ (also known as an accepting path) is a sequence of arcs from $E$ that forms a path from $q_\text{init}$ to any state in $F$.  We write $\bm z_\Sigma$, $\bm z_\Delta$, or $\bm z_\Omega$ respectively for the concatenation of the input, output, or mark strings along the arcs of $\bm z$.

The mark scoring function $\scorer: \Omega^* \to \mathbb{R}_{\geq 0}$  maps mark strings to non-negative scores.  Define 
\begin{align}
    Z_\theta\;& \defeq \ \ \ \ \ \ \ \ \ \ \sum_{\bm z \in \cal T}\ \ \ \ \ \ \ \ \ \ \tilde{p}_\theta(\bm z_\Omega) \\
    \tilde{p}_\theta(\bm x, \bm y)\; & \defeq \sum_{\bm z \in \mathcal{T}:\;\bm z_\Sigma=\bm x, \bm z_\Delta=\bm y}\tilde{p}_\theta(\bm z_\Omega)
\end{align}
where $\bm z \in \cal T$ ranges over generating paths of $\cal T$.  Provided that $Z_\theta$ is finite and positive, the NFST defines a parametric probability distribution over the generating paths of $\cal T$, namely $p_\theta(\bm z) \defeq \tilde{p}_\theta(\bm z)/Z_\theta$.   It also defines a parametric probability distribution over $\Sigma^* \times \Delta^*$, namely $p_\theta(\bm x, \bm y) \defeq \tilde{p}_\theta(\bm x, \bm y)/Z_\theta$.

\section{Parametrizing and Training NFST}\label{app::NFST_train}
Given an NFST $({\cal T}, \scorer)$ and a dataset $\cal D$ of $(\bm x, \bm y)$ pairs, a natural estimator to use for $\theta$ is the maximum likelihood estimator:
\begin{align}
    \hat{\theta}_{\mathcal{D}} &= \underset{\theta}{\argmax}\left\{ \sum_{(\bm x,\bm y)\in \mathcal{D}}[\log \tilde{p}_\theta({\bm x},{\bm y}) - \log Z_\theta
    ] \right\}
\end{align}

The difficulty is in computing $Z_\theta$ and ensuring that it is finite.  We can avoid these difficulties by dropping the $\log Z_\theta$ term.  That would be justified if $Z_\theta=1$; a first attempt to guarantee that is to adopt a locally normalized (autoregressive) model of mark strings: 
\begin{align}\label{eq:ar}
    \tilde{p}_\theta(\bm\omega) = \prod_{t=1}^{T+1} p_\theta(\omega_t \mid \bm\omega_{< t})
\end{align}
where $T=|\bm\omega|$, $\omega_{T+1}$ by convention is a distinguished end-of-sequence symbol $\textsc{eos}\notin\Omega$, and any parametric conditional probability distribution over $\Omega \cup \{\textsc{eos}\}$ may be used for the factors.  This often ensures that $\sum_{\bm\omega \in \Omega^*} \tilde{p}_\theta(\bm\omega) = 1$.  However, it does not ensure $Z_\theta = 1$ as desired.

To avoid $Z_\theta > 1$, we will require that different paths of $\cal T$ have different mark strings, so that no string in $\Omega^*$ is double-counted.  However, typically $Z_\theta < 1$ (the distribution is deficient), since some probability mass is allocated to ``ungrammatical'' mark strings that do not appear on any path of $\cal T$.\footnote{For some families of conditional distributions in \cref{eq:ar}, it is even possible for some probability mass to be allocated to infinite mark sequences \cite{du-et-al-2023}, which we also regard as ungrammatical. (Then even $\sum_{\bm\omega \in \Omega^*} \tilde{p}_\theta(\bm\omega) < 1$.)}  

By regarding the set of ungrammatical mark strings as a special event $\bot$, we may regard $\tilde{p}_\theta$ as a probability distribution over $\Omega^* \cup \{\bot\}$.   As \emph{that} distribution is already normalized ($Z_\theta + \tilde{p}_\theta(\bot) = 1$), its maximum likelihood estimator is
\begin{equation}\label{eq:mldeficient}
    \hat{\theta}_{\mathcal{D}} = \underset{\theta}{\argmax} \sum_{(\bm x,\bm y)\in \mathcal{D}}\log \tilde{p}_\theta({\bm x},{\bm y})
\end{equation}
In practice, $p_{\hat{\theta}_{\mathcal{D}}}$ will tend to place a positive but small probability on $\bot$, since $\bot$ is never observed in $\mathcal{D}$.  The maximum likelihood objective would prefer to raise the probability of grammatical mark strings and particularly the ones that can explain $\cal D$.  In other words, training $\theta$ in this way should approximately learn the grammaticality constraint $\tilde{p}_\theta(\bot) \approx 0$ rather than having it be a structural zero of the model.  In effect, it forces $\theta$ to learn something about the structure of $\cal T$ (which could, in fact, serve as a useful form of multi-task regularization that improves generalization).  

The remaining difficulty is that the log-likelihood \labelcref{eq:mldeficient} requires an intractable sum over all paths that generate $(\bm x, \bm y)$.  So instead of minimizing the negative log-likelihood, we minimize a variational upper bound (from Jensen's inequality) that can be estimated by sampling:
\begin{equation}\label{eq::loss}
\begin{aligned}
    \mathcal{L}(\theta) &\defeq -\mathbb{E}_{(\bm x,\bm y)\sim \mathcal{D}}\left[ \log \tilde{p}_\theta({\bm x},{\bm y}) \right]\\
    &=-\mathbb{E}_{(\bm x,\bm y)\sim \mathcal{D}}\left[\log \sum_{\bm z \in \mathcal{T}:\;\bm z_\Sigma=\bm x, \bm z_\Delta=\bm y} \scorer(\bm{z})\right]\\
    &= -\mathbb{E}_{(\bm x,\bm y)\sim \mathcal{D}}\left[\log \mathbb{E}_{\bm z \sim q_{\phi}} \left[\frac{\scorer(\bm z)}{q_{\phi}(\bm z \mid \bm x, \bm y)}\right]\right]\\
    & \leq -\mathbb{E}_{(\bm x,\bm y)\sim \mathcal{D}}\left[\mathbb{E}_{\bm z \sim q_{\phi}} \left[\log \frac{\scorer(\bm z)}{q_{\phi}(\bm z\mid \bm x, \bm y)}\right]\right]\\
    &\defeq \mathcal{L}'(\theta)
\end{aligned}
\end{equation}
where $q_\phi(\cdot \mid \bm x, \bm y)$ may be any conditional distribution over the generating paths of $(\bm x,\bm y)$ in $\cal T$.  The tightest bound for a given parametric family $q_\phi$ is obtained by minimizing $\mathcal{L}'(\theta)$ as a function of $\phi$, which minimizes the exclusive KL divergence \labelcref{eq:KLexc}.

For any $q_\phi$, we may use the importance weighting estimator IWAE \cite{iwae} to obtain an even tighter upper bound based on averaging over $K > 1$ paths under the log:
\begin{equation}
    \mathcal{L}'_{K}(\theta) \defeq -\mathbb{E}_{(\bm x,\bm y)\sim \mathcal{D}} \left[\underset{\bm z^{(1)}\cdots \bm z^{(K)}\sim q_{\phi}}{\mathbb{E}} \left[\log \frac{1}{K} \sum_{k=1}^K \frac{\scorer(\bm z)}{q_{\phi}(\bm z \mid \bm x, \bm y)}\right]\right]
\end{equation}
which can again be estimated by sampling.

What do we use for the autoregressive model in \cref{eq:ar}?  In our experiments, we follow \cite{lin-eisner-2018-neural} and define 
\begin{align}\label{eq:scorer}
\log p_\theta(\omega_t \mid \bm \omega_{<t}) &\propto \exp (W\,[1;\vh_{t-1}])
\end{align}
for all $\omega_t \in \Omega \cup \{\textsc{eos}\}$,
where $\vh_{t-1}$ is the state of an LSTM after reading $\bm \omega_{<t}$.  

\section{Hyperparameters}\label{app::hyperparams}

In this section, we describe the hyperparameters used for the frozen scorer and the various samplers in the experimental setup of \cref{sec::eval}.

For the frozen scorer $\scorer$, we parameterize it with a two-layer LSTM with hidden dimension 256.

For samplers $q_\phi$, the recurrent networks are one-layer.  Though the model architecture varies (as described in \cref{sec::sample}), they all use the same hidden dimension size $d$ for each of their recurrent networks (except that the bidirectional encoder for SWA uses hidden dimension $d/2$ in each direction, so that the dot product in \cref{eq:att} is conformable).  Learned embeddings for the input, output, and mark symbols also all have dimension $d$.  

\begin{figure*}[h]
    \centering
    \begin{tabular}{c}
        \includegraphics[scale=0.55, trim=0cm 0cm -2cm 0cm]{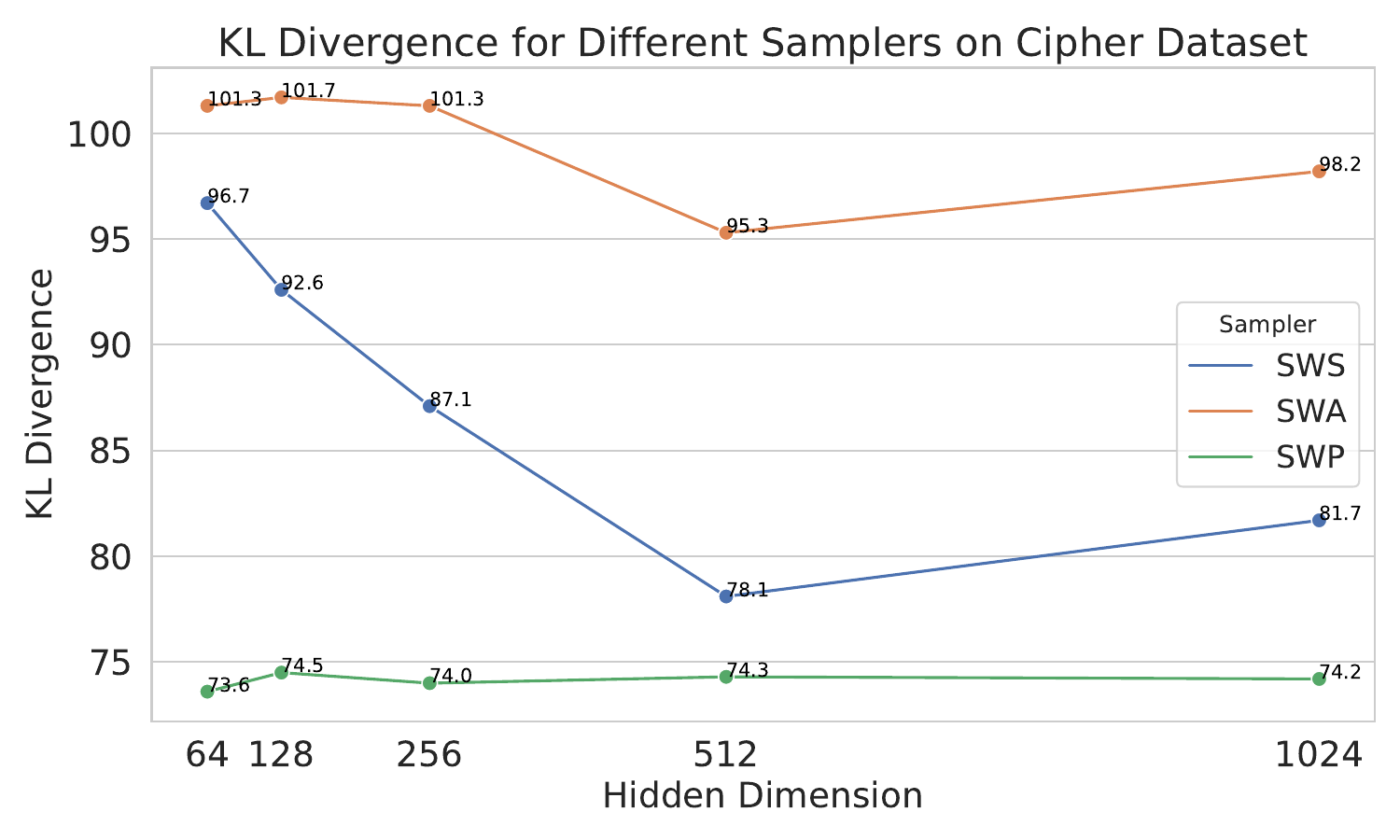} \\
    \end{tabular}
    \caption{KL divergence for different samplers on \texttt{cipher} dataset, using base models with hidden dimensions varying from 64 to 1024.
    }
    \label{fig:ablation_dim}
\end{figure*}

The result presented in \cref{table::result} is based on $d=256$. Note that while the models are matched in dimensionality, they are unfortunately not matched in the number of parameters.
We also vary the dimension size in \cref{fig:ablation_dim} to see the effect of model capacity on the samplers. The other hyperparameters used consistently for all model training are shown in \cref{table:hyperparameters}. As shown in the table, we use a small batch size (16); this is because each example in the batch samples $k=16$ proposals, leading to $16 \times 16= 256$ mark strings to generate or score in parallel. 

We also employ common techniques like length penalty for very short sequences, as well as label smoothing, gradient clipping, etc.

\begin{table}[htbp]
    \centering
    \begin{tabular}{lc}
        \toprule
        \textbf{Hyperparameter} & \textbf{Value} \\
        \midrule
        batch\_size & 16 \\
        $k$ (\# proposals) & 16 \\
        $K$ (for IWAE) & 32 \\
        dropout & 0.3 \\
        grad\_clip & 5.0 \\
        learning\_rate & $1 \times 10^{-5}$ \\
        tilde\_p\_lr & $1 \times 10^{-3}$ \\
        label\_smoothing & 0.1 \\
        length\_threshold & 100 \\
        length\_penalty & 1.0 \\
        \bottomrule
    \end{tabular}
    \caption{Hyperparameter List}
    \label{table:hyperparameters}
\end{table}

\newpage
\section{Qualitative Analysis}\label{app::qual}
In \crefrange{tab:tr_examples}{tab:cipher_examples}, we compare the mark sequences proposal by different samplers for each dataset to give a more thorough understanding of the proposal samplers' behavior. For \texttt{tr}, the input is "chalnay" and the output in Urdu is "<ur> 35 8 10 22" where <ur> is a fixed language code and the numbers denote Urdu symbols. As \texttt{tr} alignment is monotonic, we see that all three samplers give good alignments, where SWS and SWA propose the exact same path and SWP is slightly different in that \textbf{it made a mistake by aligning "c" to the language label "<ur>"}. For SWS and SWA, they align "c" to "35", "hal" to "8", "n" to "10", and "ay" to "22". For SWP, it aligns "c" to "<ur>", "hal" to "35 8", "n" to "10", and "ay" to "22". Therefore SWP is slightly wrong compared to the other two, which is also reflected in the average KL divergence in the main results (\cref{table::result}).

\texttt{scan} is a hard task for all three samplers as the alignment is not necessarily monotonic (although the example we show here is actually monotonic). We find that SWS and SWA samplers produce somewhat interpretable results (which presumably correspond to high-scoring paths under $p_\theta$), as they mix the generation of the input mark strings with the output strings. For example, SWS proposes "look around" followed by a sequence of actions, then proposes "jump around" that is followed by "I\_Jump", \etc. However, for SWP sampler, its path suggests that it did not know how to align these input and output marks because it simply dumps all the output marks first and then deletes all the input marks. As discussed in \cref{sec::experiment}, we believe this is essentially a failure of training a complex model: SWP does not learn how to usefully embed the graph of monotonic suffix alignments corresponding to paths in the deletion/insertion FST.

Lastly, for \texttt{cipher}, the SWS and SWP samplers have proposed to use cipher2, which in this case allows high-probability alignments with many copy marks that explain output symbols as enciphered versions of input symbols.  Given the choice of cipher2, SWP continues by finding a better alignment than SWS does. (SWS erred by enciphering the second "d" and the final "o" too early ("d$\rightarrow$f", "o$\rightarrow$i"), forcing it to delete those output characters and insert them again later.)
The SWA proposal has unwisely proposed cipher3, resulting in a much longer and lower-probability alignment that does not explain any of the output symbols as enciphered input symbols, but must choose to insert all of them.

\begin{table*}[h]
    \small
    \centering
    \hspace{-1cm}
    \begin{tabular}{l|c}
        \toprule
        Input & chalnay \\
        \midrule
        Output & <ur> 35 8 10 22\\
        \midrule
        SWA & insert <ur>, delete c, insert 35, delete h, delete a, delete l, insert 8, delete n, insert 10, delete a, delete y, insert 22 \\
        \midrule
        SWS & same as SWA \\
        \midrule
        SWP & delete c, insert <ur>, delete h, delete a, delete l, insert 35, insert 8, delete n, insert 10, delete a, delete y, insert 22\\
        \bottomrule
    \end{tabular}
    \caption{Proposed mark strings given an $(\bm x,\bm y)$ pair from the \texttt{tr} dataset.}\label{tab:tr_examples}

    \begin{tabular}{c|c}
        \toprule
        Input & look around left twice and jump around right \\
        \midrule
        Output & \parbox{14.2cm}{I\_TURN\_LEFT I\_LOOK I\_TURN\_LEFT I\_LOOK 
        I\_TURN\_LEFT I\_LOOK I\_TURN\_LEFT I\_LOOK I\_TURN\_LEFT 
        I\_LOOK I\_TURN\_LEFT I\_LOOK I\_TURN\_LEFT I\_LOOK I\_TURN\_LEFT I\_LOOK I\_TURN\_RIGHT I\_JUMP I\_TURN\_RIGHT I\_JUMP I\_TURN\_RIGHT I\_JUMP I\_TURN\_RIGHT I\_JUMP}\\
        \midrule
        SWA & 
        \parbox{14.2cm}{look I\_TURN\_LEFT around I\_LOOK I\_TURN\_LEFT I\_LOOK I\_TURN\_LEFT I\_LOOK I\_TURN\_LEFT I\_LOOK left I\_TURN\_LEFT I\_LOOK I\_TURN\_LEFT I\_LOOK twice I\_TURN\_LEFT I\_LOOK I\_TURN\_LEFT I\_LOOK I\_TURN\_RIGHT I\_JUMP and jump around I\_TURN\_RIGHT I\_JUMP right I\_TURN\_RIGHT I\_JUMP I\_TURN\_RIGHT I\_JUMP}
        \\
        \midrule
        SWS & \parbox{14.2cm}{look around I\_TURN\_LEFT I\_LOOK I\_TURN\_LEFT I\_LOOK I\_TURN\_LEFT left I\_LOOK I\_TURN\_LEFT twice I\_LOOK I\_TURN\_LEFT I\_LOOK I\_TURN\_LEFT I\_LOOK I\_TURN\_LEFT I\_LOOK I\_TURN\_LEFT I\_LOOK and I\_TURN\_RIGHT jump around I\_JUMP I\_TURN\_RIGHT I\_JUMP I\_TURN\_RIGHT I\_JUMP right I\_TURN\_RIGHT I\_JUMP}\\
        \midrule
        SWP & \parbox{14.2cm}{I\_TURN\_LEFT I\_LOOK I\_TURN\_LEFT I\_LOOK 
        I\_TURN\_LEFT I\_LOOK I\_TURN\_LEFT I\_LOOK I\_TURN\_LEFT 
        I\_LOOK I\_TURN\_LEFT I\_LOOK I\_TURN\_LEFT I\_LOOK look I\_TURN\_LEFT I\_LOOK I\_TURN\_RIGHT I\_JUMP I\_TURN\_RIGHT I\_JUMP I\_TURN\_RIGHT I\_JUMP I\_TURN\_RIGHT I\_JUMP around left twice and jump around}\\
        \bottomrule
    \end{tabular}
    \caption{Proposed mark strings given an $(\bm x,\bm y)$ pair from the \texttt{scan} dataset.}\label{tab:scan_examples}
    \begin{tabular}{c|c}
        \toprule
        Input &d d o r g r r i q a v o\\
        \midrule
        Output &f j f i a e a a j k s z w i e\\
        \midrule
        SWA & \parbox{14.2cm}{cipher3: d$\rightarrow$m$\rightarrow$$\varepsilon$, insert f, insert j, d$\rightarrow$m$\rightarrow$$\varepsilon$, insert f, insert i, o$\rightarrow$x$\rightarrow$$\varepsilon$, insert a, insert e, r$\rightarrow$f$\rightarrow$$\varepsilon$, insert a, insert a, g$\rightarrow$z$\rightarrow$$\varepsilon$, insert j, insert k, r$\rightarrow$f$\rightarrow$$\varepsilon$, insert s, insert z, r$\rightarrow$f$\rightarrow$$\varepsilon$, insert w, insert i, i$\rightarrow$j$\rightarrow$$\varepsilon$, insert e, q$\rightarrow$a$\rightarrow$$\varepsilon$, a$\rightarrow$e$\rightarrow$$\varepsilon$, v$\rightarrow$h$\rightarrow$$\varepsilon$, o$\rightarrow$x$\rightarrow$$\varepsilon$}\\
        \midrule
        SWS & \parbox{14.2cm}{cipher2, d$\rightarrow$f, d$\rightarrow$f$\rightarrow$$\varepsilon$,  insert j, insert f, o$\rightarrow$i, r$\rightarrow$a, g$\rightarrow$e, r$\rightarrow$a, r$\rightarrow$a, i$\rightarrow$j, q$\rightarrow$k, a$\rightarrow$s, v$\rightarrow$z, o$\rightarrow$i$\rightarrow$$\varepsilon$, insert w, insert i, insert e} \\
        \midrule
        SWP & \parbox{14.2cm}{cipher2: d$\rightarrow$f, insert j, d$\rightarrow$f, o$\rightarrow$i, r$\rightarrow$a, g$\rightarrow$e, r$\rightarrow$a, r$\rightarrow$a, i$\rightarrow$j, q$\rightarrow$k, a$\rightarrow$s, v$\rightarrow$z, insert w, o$\rightarrow$i, insert e}\\
        \bottomrule
    \end{tabular}
    \caption{Proposed mark strings given an $(\bm x,\bm y)$ pair from the \texttt{cipher} dataset.  The notation "a$\rightarrow$b" is actually an abbreviation for a sequence of 5 marks: replace, a, b, copy, b. 
    The notation "a$\rightarrow$b$\rightarrow$$\varepsilon$" is also an abbreviation for a sequence of 5 marks: replace, a, b, delete, b.
}
    \label{tab:cipher_examples}
\end{table*}

\begin{table*}[!ht]
    \centering
    \begin{tabular}{cccc}
      \toprule
        Task & TR & Scan & Cipher  \\ \midrule
        SWA & 1.0 & 7.9  & 1.1  \\ 
        SWS & 1.1 & 4.3 & 1.3  \\ 
        SWP & 1.0 & 1.1 & 1.0 \\ \bottomrule
    \end{tabular}
    \caption{Deduplicated effective sample size (ESS) from different samplers across tasks. Instead of using the regular ESS, we first aggregate the probability mass for samples that are identical as we realize that many duplicated samples are proposed. Then we compute the normalized probability $\hat{w}$ over the deduplicated probabilities and compute the $\text{ESS} = \frac{1}{\hat{w}^2} = \frac{(\sum_{i=1}^n w_i)^2}{\sum_{i=1}^n w_i^2}$. We see that ESS is very small for \texttt{tr} and \texttt{cipher} task, showing that the proposal distribution is sharp after training.}
    \label{table::ess}
\end{table*}

\section{Baseline Sampler with No Lookahead}\label{app::no_lookahead}
We trained a baseline sampler that does not utilize information from the future (suffix of input $\bm x$ and output $\bm y$). The parameterization is shown below:
\begin{equation}
    q_\phi(\omega_t \mid \bm \omega_{<t}, \Txy)  
        = \text{softmax} (W [1; \vh_{t-1}])
\end{equation} 

We would expect the samplers with lookahead to give better performance compared to this no-lookahead baseline, but our result in \cref{table::no_lookahead} shows that SWP performs worse compared to it on \texttt{tr} and \texttt{scan} tasks. This suggests that SWP might suffer from training issues or that even a well-trained model with the SWP architecture tends to generalize poorly.

\begin{figure}[ht]
\centering
\begin{tabular}{cccccc}
    \toprule
    Sampler & TR & Scan & Cipher & Avg \\
    \midrule
    SWA & 23.3 & 38.1 & 101.3 & 54.3 \\
    SWS & \textbf{21.8} & \textbf{25.4} & 87.1 & \textbf{44.7} \\
    SWP & 38.1 & 102.9 &  \textbf{73.9} & 71.7 \\
    No Lookahead & 31.9 & 39.9 & 95.5&55.8  \\
    \bottomrule
\end{tabular}
\caption{\textit{Partial KL} divergence comparison (lower values are better) for different proposal samplers.}
\label{table::no_lookahead}
\end{figure}

\newpage
\section{Limitations}\label{app::limitation}

\paragraph{Inference Speed}
In this work, we mainly focus on the quality aspect of inference networks, rather than their speed. However, many applications of inference networks depend on their speed \cite{Leviathan2022FastIF}; otherwise, comparable quality might be achieved using a more naive sampling strategy. For our samplers, SWA is the fastest because it directly computes attention over the representation of the raw strings.  SWS is only slightly slower.  Our novel method, SWP, is the slowest (about 10x time of SWA) as it propagates information over the large graph of possible alignments defined by the FST topology.

\paragraph{Cyclic MFSTs} Our SWP sampler assumes that the MFST $\Txy$ is acyclic.  Since $\Txy$ is deterministic and hence $\varepsilon$-free with respect to the mark tape, this is equivalent to saying that for each $\bm x, \bm y$, the length of a mark string is bounded.  This assumption is ordinarily satisfied, but might be violated if, for example, $\cal T$ were obtained by composing an MFST that can insert arbitrarily many symbols with another MFST that can delete them again.  This would lead to infinitely many generating paths that transform $\bm x$ into $\bm y$, which would require a cyclic $\Txy$ encode.

\paragraph{Larger Datasets and Models} Due to the computation cost of the parameter estimation that requires sampling and state tracking, we are using three datasets with small vocabulary to showcase the difference between samplers. We will leave scaling up the dataset/model for future investigation.

\paragraph{Alternative Architectures} We used recurrent neural networks as our main building block. Switching to Transformers, for example, would require a redesign of the SWP sampler to attend over arcs in the suffix graph.  We also limited our experiments to single-layer recurrent networks.

\paragraph{Particle Smoothing} Due to computation cost, we also did not perform full particle smoothing (used in \cite{lin-eisner-2018-neural}) because it requires sampling of multiple paths $\bm z$ in parallel and weighting the prefixes $\bm z_{< t}$ \emph{at each time step $t$} by an estimate of their conditional probability. The benefit of calculating these weights at each stage lies in their ability to measure effective sample size (ESS) and facilitate multinomial resampling when the ESS drops below a certain threshold.  As shown in \cref{table::ess}, ESS indeed falls low as many duplicated paths are proposed.

\end{document}